\documentclass[letterpaper, 10 pt, conference]{ieeeconf}  

\IEEEoverridecommandlockouts                              

\usepackage{bm}
\usepackage{amsmath,graphicx}
\usepackage{amssymb}
\usepackage[colorlinks, linkcolor=black,  citecolor=black, urlcolor=blue]{hyperref}
\usepackage{multirow}

\title{\LARGE \bf
Fast Diffeomorphic Image Registration using Patch based Fully Convolutional Networks
}

\author{Jiong Wu$^{1,*}$, Shuang Zhou$^{2}$, Li Lin $^{3}$, Xin Wang$^{1}$ and Wenxue Tan$^{1}$ 
	\thanks{*This study was supported by the National Natural Science Foundation of China (62206093), the Natural Science Foundation of Hunan Province (2022JJ40290, 2022JJ50253), the Youth Foundation of Hunan Province Department of Education (21B0619) and the Scientific Research Project of Hunan University of Arts and Science (20ZD01).}
	\thanks{*Correspondence at {\tt\small wujiong@huas.edu.cn}}
	\thanks{$^{1}$ School of Computer and Electrical Engineering, Hunan University of Arts and Science, Hunan, China.}
	\thanks{$^{2}$ Furong College, Hunan University of Arts and Science, Hunan, China.}
	\thanks{$^{3}$ Department of Electrical and Electronic Engineering, The University of Hong Kong, Hongkong, China}
}

\begin{document}

\maketitle
\thispagestyle{empty}
\pagestyle{empty}

\begin{abstract}
Diffeomorphic image registration is a fundamental step in medical image analysis, owing to its capability to ensure the invertibility of transformations and preservation of topology. Currently, unsupervised learning-based registration techniques primarily extract features at the image level, potentially limiting their efficacy. This paper proposes a novel unsupervised learning-based fully convolutional network (FCN) framework for fast diffeomorphic image registration, emphasizing feature acquisition at the image patch level. Furthermore, a novel differential operator is introduced and integrated into the FCN architecture for parameter learning. Experiments are conducted on three distinct T1-weighted magnetic resonance imaging (T1w MRI) datasets. Comparative analyses with three state-of-the-art diffeomorphic image registration approaches including a typical conventional registration algorithm and two representative unsupervised learning-based methods, reveal that the proposed method exhibits superior performance in both registration accuracy and topology preservation.


\end{abstract}

\section{INTRODUCTION}
Diffeomorphic image registration plays a critical role in medical image analysis tasks such as atlas-based image segmentation \cite{wu2021brain,wu2019joint,wu2023multi} and image fusion \cite{wang2023mse}. Traditional diffeomorphic image registration methods are generally based on some energy models which convert the medical image registration problem into an optimization one by defining an energy-related objective function \cite{avants2008symmetric, wu2020large}. Iterative non-linear optimization algorithms are adopted to minimize the objective function to generate diffeomorphic transformation, which results in high accuracy but expensive computation.

With the development of deep learning techniques in computer vision, fully convolutional networks (FCNs) that facilitate voxel-to-voxel learning are successfully applicated in medical image segmentation tasks \cite{ronneberger2015u}, they are also directly adopted in medical image registration \cite{dalca2018unsupervised}. Since just one forward calculation through the trained FCNs during the registration inference, the FCN-based methods are much faster than the traditional ones.  Such methods can be divided into two categories including supervised learning-based methods and unsupervised learning-based methods. 

In supervised learning-based methods, FCNs learn maps between inputted image pairs and specifical outputs in a data-driven manner, the purpose of the networks optimization process is to minimize the difference between the predicted outputs and ground truth displacement fields \cite{sokooti2017nonrigid} or velocity fields \cite{yang2017quicksilver}, which are typically challenging to obtain in clinical practice, resulting in the limited application of these methods.  To tackle this problem, many unsupervised learning-based approaches are proposed. These methods predict displacement fields by just feeding FCNs with image pairs without any ground truth information \cite{balakrishnan2018unsupervised}. Although these methods achieved some promising results, FCNs capture the features from the whole image level, which may limit the registration accuracy since local information is not fully captured.


To achieve diffeomorphic image registration, many studies built the registration model on a stationary velocity field system. The final diffeomorphism is calculated by integrating the predicted velocity field in unit time. For instance, in \cite{dalca2018unsupervised} a probabilistic generative model was adopted to predict the stationary velocity field via the mean and covariance of the posterior registration probability. The L2 norm of spatial gradients of the velocity field is introduced into the loss function to guide the FCNs to learn a smooth stationary velocity field \cite{mok2020large}. To further improve the diffeomorphic property of generated transformation, some other studies introduced a Jacobian determinant regularization acting on the displacement field \cite{mok2020fast}. In addition, several studies utilized multi-resolution strategies to capture large deformation between image pairs resulting in accuracy improvement and desirable diffeomorphic properties maintaining \cite{wu2024diffeomorphic}.  On the other hand, \cite{wu2023hnas} proposed a neural network architecture searching algorithm to find an optimal FCN for image accurate registration. All these approaches learning the velocity field from the whole inputted image pairs still have the aforementioned limitations. In addition, although L2 norm regularization and Jacobian determinant largely improve the diffeomorphic property for the transformations, the accuracy of the registration is limited.  

In this paper, to address the limitations of the aforementioned methods, we present a novel image patch-based registration framework that calculates the velocity field by focusing on the information of the local image patch. More importantly, drawing inspiration from the conventional large deformation diffeomorphic metric mapping (LDDMM) method \cite{wu2020large}, we incorporate a differential operator into the proposed registration approach. This operator is then applied to the velocity field, aiming to enhance registration performance while preserving desired diffeomorphic properties.

\begin{figure*}[!pt]
	\includegraphics[width=18cm]{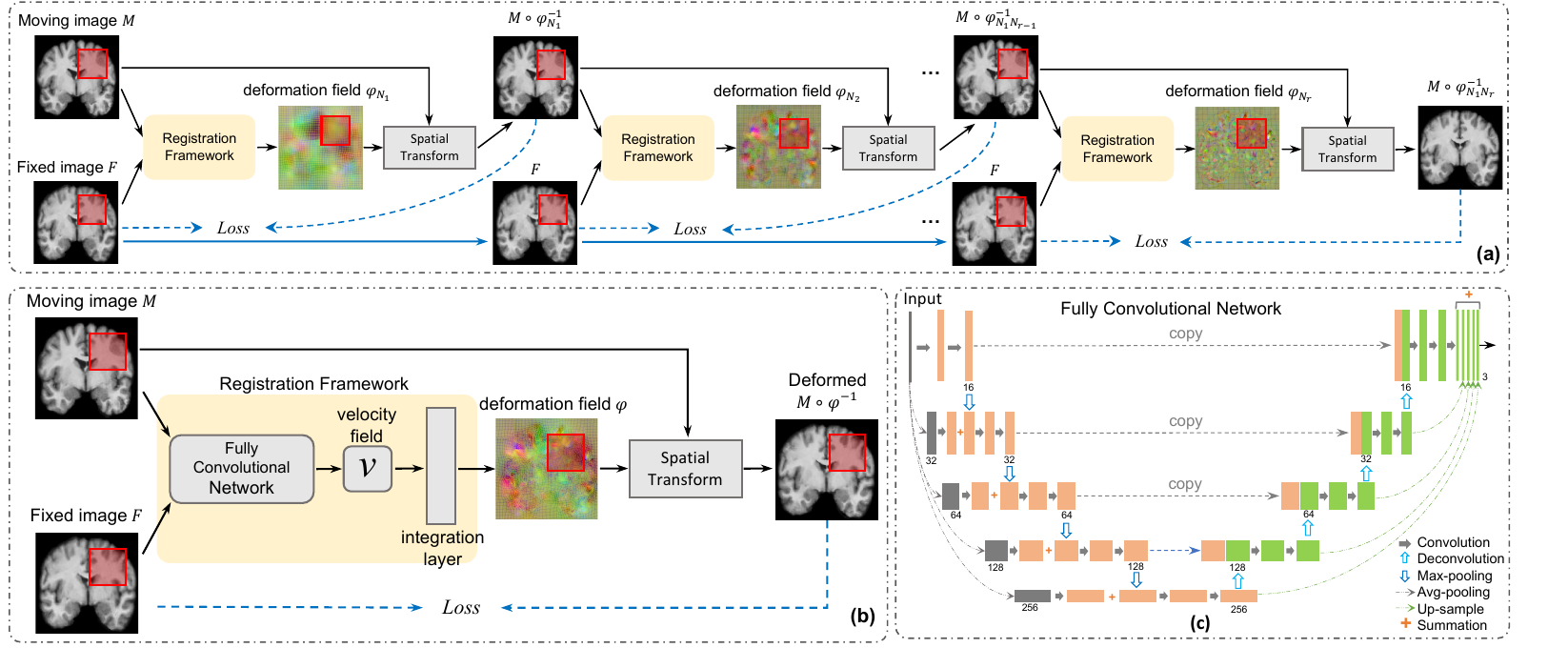}
	\centering
	\caption{(a) The architecture of the proposed coarse-to-fine patch based diffeomorphic registration framework. (b) Illustration of the sub-registration framework. (c) The architecture of the employed fully convolutional network (FCN) in our method.} 	\label{fig1}
\end{figure*}

\section{Methods}
\subsection{Patch based Diffeomorphic Registration}
Given a 3D grayscale moving image $M$ and a 3D grayscale fixed image $F$ defined on the background space $\Omega\in{\mathbb{R}^3}$. The proposed patch based diffeomorphic registration framework is to find the optimal diffeomorphism patches $\{\varphi_i|i=1,\cdots,{Q}\}$ for all image patch pairs (contain moving image patches and corresponding fixed image patches), which is subsequently integrated to generate an optimal diffeomorphic transformation $\varphi$. This transformation ensures that $M \circ \varphi^{-1}$ aligns well with $F$. In our framework, the $i$th diffeomorphism transformation patch $\varphi_i$ is defined via the following ordinary differential equation (ODE) 
\begin{equation}
	\label{equ:fov1}
	\frac{d\phi_i^t}{d{t}}=v_i^t(\phi_i^t), 
\end{equation}
where $\phi_i^0=id$ denotes the identity mapping such that $id(x)=x$, $x\in{\Omega}$, $v_i^t:\Omega\times{t}\rightarrow{\mathbb{R}^3}$ denotes stationary velocity field of the $i$th patch. After integrating the ODE over $t=[0,1]$, the final diffeomorphic transformation $\varphi_i$ can be obtained. 

\subsection{Differential Operator Embedding}
Let $\mathbf{\theta}$ be the parameters of FCN, existing learning-based diffeomorphic registration methods train the FCN by minimizing the value of loss function $\mathcal{L}$, which has the following form
\begin{equation}
	\label{equ:fov2} 
	\mathcal{L}(\theta)= \mathcal{L}_{sim}(M_i\circ{\varphi_i^{-1}}, F_i)+\lambda{\mathcal{L}_{smooth}}(v_i),
\end{equation} 
where $M_i$ and $F_i$ denote the $i$th moving and fixed image patch, $\mathcal{L}_{sim}(\cdot,\cdot)$ denotes a matching term that ensures the warped moving image patch is similar to the fixed image patch, $\mathcal{L}_{smooth}(\cdot)$ denotes a regularization term used to evaluate the smoothness of $v_i$, and $\lambda$ is a hyperparameter to balance the image matching and velocity smoothness terms. 

In unsupervised learning-based diffeomorphic registration frameworks, the L2-norm of spatial gradients of the velocity field is usually adopted which has the form of $\|\nabla{v}\|_2^2$. Different from these methods,  we embed a differential operator $L=-\alpha{\nabla}^2+Id$ into the loss function $\mathcal{L}(\theta)$ of our FCN, with $\nabla^{2}$ and $Id$ respectively denote the Laplacian and identity operators, and the hyperparameter $\alpha>0$ determines the smoothness of $v_i$. Therefore, the form of  $\mathcal{L}(\theta)$ can be derived as 
\begin{equation}
	\label{equ:fov3} 
	\mathcal{L}(\theta)=\mathcal{L}_{sim}(M_i\circ{\varphi_i^{-1}}, F_i)+\lambda\sum_{j=1}^{N}{\Vert{Lv_i^t}\Vert}_{L^2}^2\delta{t},
\end{equation}  
where $N$ denote the number of steps for velocity integration and $\delta{t}=\frac{1}{N}$. $L$ is a symmetric and positive-definite differential operator that maps a tangent vector $v_i\in{V}$ ($V$ is a tangent space of diffeomorphisms) into the dual space $V^{*}$ resulting in the calculation of $\mathcal{L}(\theta)$ is conducted in the dual space $V^{*}$. Therefore, the gradient of the loss function ${\nabla}_{\theta}{\mathcal{L}}$ is transformed to the original space to realize the backpropagation and is computed as 
\begin{equation}
	\label{equ:fov4}
	{\nabla}_{\theta}\mathcal{L}=-\frac{1}{2\sigma^2}K\left(\partial{\mathcal{L}_{sim}}(J_i, F_i)\cdot\nabla{J_i}\right)+v_i,
\end{equation}
where $J_i=M_i\circ{\varphi_i}^{-1}$, $K=(L^{\dag}{L})^{-1}$, $L^{\dag}$ denotes the adjoint of $L$,  $\nabla{J_i}$ denotes the gradient of $J_i$, $\partial{\mathcal{L}_{sim}}(J_i, F_i)$ denotes the Gateaux derivative of $\mathcal{L}_{sim}(J_i, F_i)$. 
In this paper, we set the $\mathcal{L}_{sim}$ to be the negative of NCC.

\subsection{Differential Operator based Coarse-to-fine Diffeomorphic Registration framework}
After introducing the differential operator $L$, we can note that the different values of $\alpha$ will generate different smoothness of velocity fields (the smaller of the $\alpha$ the more smoothness of the velocity field) \cite{wu2020large}. Therefore, to handle large deformation between image pairs, we adopt a coarse-to-fine image registration framework with gradually decreasing values of $\alpha$ to progressively estimate the velocity and deformation fields.
Fig. 1 (a) and (b) illustrate the overall framework of the proposed diffeomorphic image registration method involving $N_r$ different stages. Specifically, the velocity fields are estimated with incremental smoothness in $N_r$ levels ($k=N_1$ and $k=N_r$ denote the coarsest and finest smoothness, respectively). The input of the $N_k$th FCN is the warped moving image 
$M\circ\varphi_{N_1N_{k-1}}^{-1}$ and fixed image $F$ with $\varphi_{N_1N_{k-1}}=\varphi_{N_{i-1}}\cdot\varphi_{N_1N_{k-2}}$. Finally, we can obtain the deformation $\varphi_{N_1N_r}$ by combining $N_r$ of $\varphi$ together.

The proposed FCN has an encoder-decoder structure with skip connections, building on top of the 3D UNet-style architecture \cite{ronneberger2015u}, as depicted in Fig. \ref{fig1} (c). Specifically, we adopt 3D convolutions followed by Leaky ReLU activations in each layer of both the encoder and decoder parts. Small kernels of sizes $3\times3\times3$ and $2\times2\times2$ are used alternatively. To fully capture the hierarchical features of the input image pair, in addition to two traditional down-sampling and up-sampling streams, we build another eight paths with four for contracting and four for expanding. Specifically, four times of average pooling followed by a convolutional layer are applied to the input image pair and added to the feature maps that have the same size as the encoder part.  In the decoder, we use trilinear interpolation to up-sample four different sizes of feature maps and output four full-scale feature maps each having 3 channels. In the last layer of our FCN, five different outputs are added together as the finally velocity field.

\subsection{Implementation Details}
In this study, we adopt a semi-Lagrangian scheme \cite{wu2020large} to calculate the transformation. 
FCNs and calculations of the operators $L$ and $K$ are directly implemented using Pytorch \footnote{\url{https://pytorch.org/}}, and all other components of the loss function $\mathcal{L}(\theta)$ (Eq. \ref{equ:fov3}) and all components of the derivative $\nabla_{\theta}{\mathcal{L}}$  (Eq. \ref{equ:fov4}), including the NCC, the Gateaux derivative of NCC, the Jacobian determination of $\varphi_i$, the semi-Lagrangian scheme and the spatial transformation module, are firstly implemented using CUDA \footnote{\url{https://docs.nvidia.com/cuda/}}, and then integrated into the Pytorch framework. 

To balance registration performance and large displacements, we adopt a three-stage coarse-to-fine registration flow ($N_r=3$). And the final optimal spatial transformation $\varphi$ is calculated by combining the diffeomorphisms $\varphi_1$, $\varphi_2$, and $\varphi_3$ obtained from the three cascading registration stages. Three decreasing $\alpha$ values (0.01, 0.005, 0.001 respectively) are set in these three stages, $\sigma$ is set to be $0.001$ and the length of the local volume used to compute the NCC is set to be 7. In the training process, our patch-based networks are trained by the fixed size of image patch pairs, and a Dense training scheme \cite{wu2019joint} is adopted to equally sample the patch pairs from foreground and background. The size of the training patches is set to be $64\times64\times64$ and the batch size is set to be 16. In the testing procedure, velocity field vectors only located in a cubic (with the size of $32\times32\times32$) at the center of each patch are considered. The parameter optimization in the FCN is performed using $Adam$ with a learning rate of $1\times{10^{-4}}$. For training the current level of FCN, parameters of previous levels of FCNs are fixed. 

\section{Experimental Results and Evaluation}
\subsection{Datasets and Evaluation Metrics}
To evaluate the proposed registration method, we adopted two datasets including (a) Mindboggle-101 dataset \footnote{\url{https://mindboggle.info/data\#}} consisting of 101 T1-weighted images with 50 manually delineated cortical structures, and (b) MICCAI 2012 Multi-Atlas Labelling Challenge (MALC) dataset \footnote{\url{http://www.neuromorphometrics.com/2012_MICCAI_Challenge_Data.html}} comprising 35 T1-weighted images with 134 manually delineated brain structures as the testing datasets. For training the framework, 370 T1-weighted images were randomly selected from the OASIS dataset \cite{marcus2007open}, 362 images made up of 182 image pairs for training, and 6 images made up of 5 pairs for validation. Skull-stripping and brain regions automatically segmented on the OASIS dataset were conducted by using FreeSurfer \cite{fischl2012freesurfer}. Moreover, we used FreeSurfer to conduct 12 parameters of affine spatial normalization on these datasets. Then intensity normalization between 0 and 1, spatial resampling with the resolution of $1\times1\times1$ $mm^3$, and center cropping with the size of  $160\times192\times224$ were conducted. 

We evaluate our method using two measurements, including the Avg. DSC is calculated by averaging the DSCs across all structures in segmentation maps along with the corresponding average standard deviation, and the number and percentage of voxels with non-positive Jacobian determinants as well as corresponding standard deviation.

\begin{table*}[t]
	\centering
	\caption{Average Dice score, number and percentage of voxels with non-positive Jacobian determinant for affine alignment, SyN, Diff-VM, SYMNet, and the proposed method on different testing datasets}
	\resizebox{\textwidth}{!}{
	\footnotesize
	\begin{tabular}{c|ccc|ccc}
		\hline
		\multirow{2}{*}{Methods} & \multicolumn{3}{c|}{MICCAI2012}                                                                       & \multicolumn{3}{c}{Mindboggle101}                                                                    \\ \cline{2-7} 
		& \multicolumn{1}{c|}{Avg.DSC}       & \multicolumn{1}{c|}{$|J\phi|\leq0$} & $|J\phi|\leq0$ (\%) & \multicolumn{1}{c|}{Avg.DSC}       & \multicolumn{1}{c|}{$|J\phi|\leq0$} & $|J\phi|\leq0$ (\%) \\ \hline
		Affine                   & \multicolumn{1}{c|}{0.438 (0.058)} & \multicolumn{1}{c|}{$-$}                     &           $-$           & \multicolumn{1}{c|}{0.354 (0.017)} & \multicolumn{1}{c|}{$-$}                     &     $-$                 \\ 
		SyN                   & \multicolumn{1}{c|}{0.588 (0.031)} & \multicolumn{1}{c|}{2953 (1116)}          & 5.10e-2 (1.60e-2)      & \multicolumn{1}{c|}{0.534 (0.019)}  & \multicolumn{1}{c|}{1275 (675)}            & 1.85e-3 (9.82e-3)    \\ 
		Diff-VM                  & \multicolumn{1}{c|}{0.580 (0.032)} & \multicolumn{1}{c|}{27.9 (16.93)}                & 4.05e-4 (4.23e-4)                & \multicolumn{1}{c|}{0.515 (0.023)} & \multicolumn{1}{c|}{42.7 (21.65)}                & 6.21e-4 (3.27e-4)                \\ 
		SYMNet                   & \multicolumn{1}{c|}{0.585 (0.031)} & \multicolumn{1}{c|}{36.7 (22.16)}         & 5.30e-4 (5.34e-4)     & \multicolumn{1}{c|}{0.547 (0.020)} & \multicolumn{1}{c|}{46.7 (22.87)}         & 6.79e-4 (3.32e-4)    \\ 
		Our method                & \multicolumn{1}{c|}{0.591 (0.032)} & \multicolumn{1}{c|}{37.5 (79.13)}         & 5.45e-4 (5.46e-4)    & \multicolumn{1}{c|}{0.554 (0.018)} & \multicolumn{1}{c|}{44.4 (22.33)}        & 6.45e-4 (3.25e-4)    \\ \hline
	\end{tabular}
}
\end{table*}

\subsection{Results and Discussion}
To present the efficacy of our approach, we conducted comparative experiments with three typical representative registration algorithms. These included a traditional diffeomorphic registration method known as symmetric diffeomorphic image registration (SyN) \cite{avants2008symmetric}, along with two unsupervised learning-based methods, namely DIF-VM \cite{dalca2018unsupervised} and SYMNets \cite{mok2020fast}. The experimental settings are the same as those in these three studies. NCC was used as the image similarity regularization in these three approaches. Testing experiments were respectively performed on MACL and Mindboggle101 datasets. One image of each dataset was randomly selected as the fixed image, and the remaining 34 and 100 images were considered as the moving images. 

Table 1 illustrates the average DSC, number, and percentage of voxels with non-positive Jacobian determinants for the affine normalization, SyN, Diff-VM, and the proposed method. Our method outperforms the other three diffeomorphic approaches regarding registration accuracies for these two datasets while maintaining the number of voxels with a non-positive Jacobian determinant close to zero. Although both the number and percentage of voxels with non-positive Jacobian determinants of Diff-VM are the smallest, registration accuracies of our proposed method with 1.1\% and 3.9\% higher than those of Diff-VM.  Fig. 2 shows an example of axial MR slices of moving image, fixed image, and warped image from SYMNet, SyN, Diff-VM, and our method. We can observe that the deformed result from our proposed framework is closest to the fixed image.  The average running time of SyN, Diff-VM, SYMNet, and our method is about 1039, 0.517, 0.414, and 1.51 seconds, respectively, suggesting that our method achieves superior registration accuracy in a promising registration speed. 

\begin{figure}[!pt]
	\includegraphics[width=8.0cm]{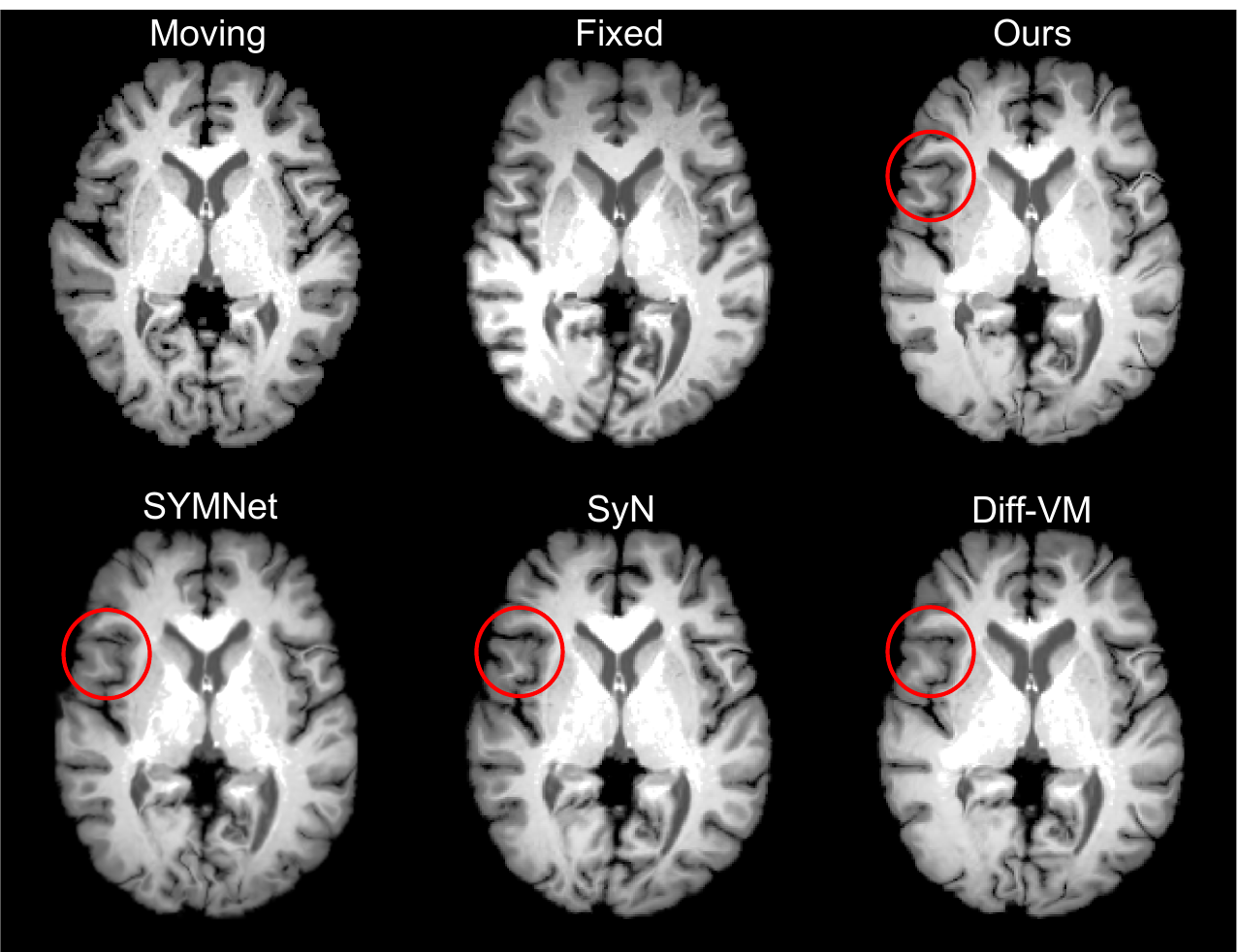}
	\centering
	\caption{Example MR slices of moving image, fixed image, and resulting deformed image from our method, SYMNet, SyN, and Diff-VM. Red circles denote the differences across different methods.} 	\label{fig2}
\end{figure}

\section{CONCLUSIONS}

We proposed and validated a novel unsupervised learning-based FCN framework for fast diffeomorphic image registration which conducts the learning and prediction in an image patch manner. In addition, after the embedding of a differential operator, a new coarse-to-fine registration framework based on different smoothness of the predicted velocity fields is proposed to capture large deformations of image pairs. Experimental results on two datasets with a different number of brain structures demonstrated that our proposed method performed significantly better while preserving desirable diffeomorphic properties and promising registration speed compared to other state-of-the-art methods.

\addtolength{\textheight}{-12cm} 

\bibliographystyle{IEEEtran}
\bibliography{IEEEabrv,refs}

\end{document}